\documentclass[runningheads]{llncs}

\usepackage[pdftex]{graphicx}
  \graphicspath{{images/}}
  \DeclareGraphicsExtensions{.pdf,.jpeg,.png,.jpg}
\usepackage{listings} 
\usepackage{etoolbox}
\apptocmd{\thebibliography}{\raggedright}{}{}

\begin{document}

\title{Hierarchical ResNeXt Models for Breast Cancer Histology Image Classification}
  \author{Isma\"el Kon\'e\footnote{Corresponding author: \email{i.kone@edu.umi.ac.ma}. \\ 
  This article should be cited as: \textit{Kon\'e I., Boulmane L. (2018) Hierarchical ResNeXt Models for Breast Cancer Histology Image Classification. In: Campilho A., Karray F., ter Haar Romeny B. (eds) Image Analysis and Recognition. ICIAR 2018. Lecture Notes in Computer Science, vol 10882. Springer, Cham}. } \and Lahsen Boulmane }
\authorrunning{Isma\"el K. and Lahsen B.} 
%
\tocauthor{Isma\"el Kon\'e, Lahsen Boulmane}
\institute{2MIA Research Group, LEM2A Lab, Facult\'e des Sciences, 
           Universit\'e Moulay Ismail, Meknes, Morocco}

\maketitle              

\begin{abstract}
Microscopic histology image analysis is a cornerstone in early detection of breast cancer.
However these images are very large and manual analysis is error prone and very time consuming.
Thus automating this process is in high demand. We proposed a hierarchical system of convolutional 
neural networks (CNN) that classifies automatically patches of these images into four pathologies: normal, benign, in situ carcinoma and invasive carcinoma. We evaluated our system on the BACH challenge dataset of image-wise classification and a small dataset that we used to extend it. Using a train/test split of 75\%/25\%, we achieved an accuracy rate of 0.99 on the test split for the BACH dataset and 0.96 on that of the extension. On the test of the BACH challenge, we've reached an accuracy of 0.81 which rank us to the $8^{th}$ out of 51 teams.
\end{abstract}

\section{Introduction}
\label{probdef}

Every year, breast cancer kills more than 500,000 women around the world \cite{who}. Early detection can help take proper actions before the spread of cancerous tissues. This has been proved to reduce death rate in US \cite{early}. Histology image analysis is necessary to perform early diagnosis \cite{acs}. However these images are too large so manually analyzing them is very time consuming and error prone. Thus an automated system is more than welcome to reduce the burden of manual analysis. 

In this scope, the BreAst Cancer Histology (BACH) Challenge\footnote{https://iciar2018-challenge.grand-challenge.org/dataset/} has been organized to stimulate scientific interest in solving this problem and finding a solution that will be a step closer to the equipment of clinical centers with such system. Specifically, it is about classifying histology images in four pathological groups: \textit{Normal}, \textit{Benign}, \textit{In situ carcinoma} and \textit{Invasive carcinoma}.

Our main contribution is a hierarchy of Convolutional Neural Network (CNN) models that gradually classify images from general pathological groups namely carcinoma and non-carcinoma and then into the four groups cited above. We trained our system on 75\% of the challenge dataset (Sect.~\ref{train}) and evaluated it on the remaining 25\% (Sect.~\ref{results}). We also extended this dataset with another one and performed the same training and evaluation.

\begin{figure}[!t]
\centering
\includegraphics[width=4.5in]{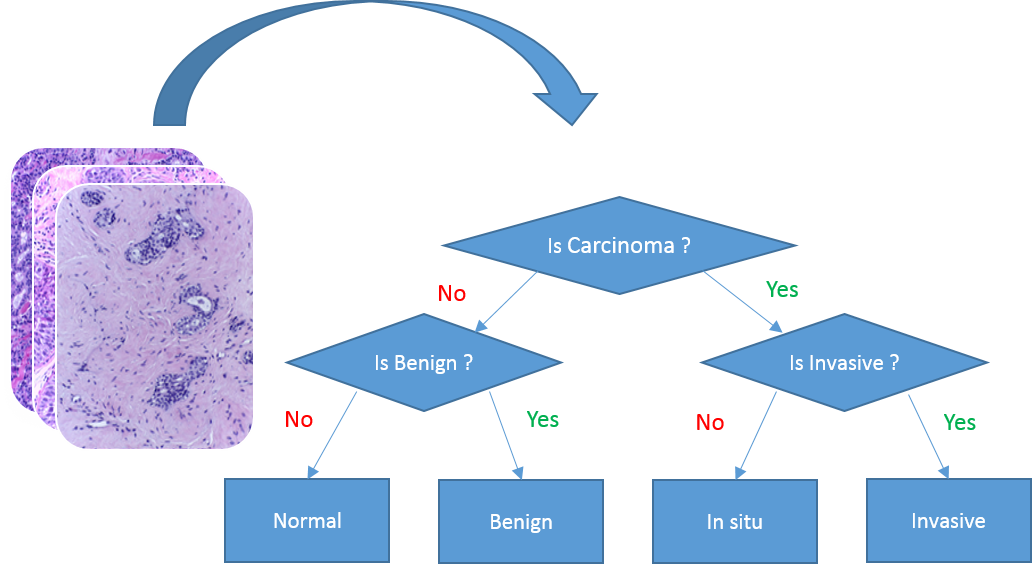}
\caption{Binary tree structure of the hierarchy of models for classifying images into the four pathological groups. Each question is answered by a specific model. The model in charge of classifying images into Carcinoma and Non Carcinoma deals with the top question. For the remaining questions, their models classify images into their respective children in the tree structure.}
\label{model}
\end{figure}

\section{Method}
\label{method}

Instead of classifying images directly into the four pathological groups that may be difficult to differentiate, our approach is about starting by a simplified version of the problem that is classifying images into two categories:

\begin{itemize}
  \item Carcinoma: which includes In Situ and Invasive pathologies.
  \item Non Carcinoma: which includes Normal and Benign pathologies.
\end{itemize}
Then we classify images from each category into the two pathologies that composed them.
We use a CNN model for each classification.
So a CNN model that we call the general model is in charge of this simplified version of the problem.
Then we have two specialized CNN models that classify respectively the Carcinoma category into In Situ and Invasive and the Non Carcinoma category into  Normal and Benign. Thus we have a hierarchy of three CNN models in a binary tree structure  where each model/node classifies incoming images into two types. \figurename ~\ref{model} synthesizes visually the structure of the hierarchy of models where each one answers a question.
We describe below the underlying CNN model used and how we train them.

\subsection{Model}

\subsubsection{Architecture:}
\begin{figure}[!t]
  \centering
  \includegraphics[width=\textwidth]{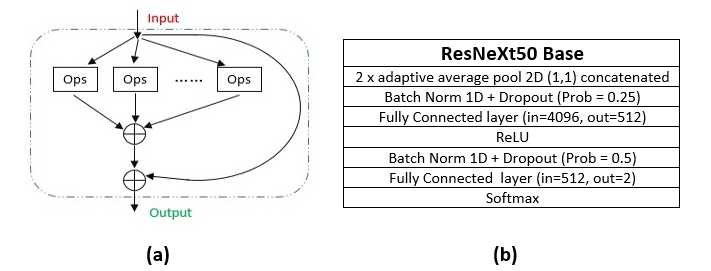}
  \caption{(a)ResNeXt block. Ops stands for convolution and non-linear operations. 
           (b)ResNeXt50 adapted to our task. ResNeXt50 Base is the original ResNeXt50 without its last layer. Below are custom last layers we added for our classification.}
  \label{resnext}
\end{figure}

For all three models, we used the ResNeXt50 architecture \cite{resnext} which is structured in repetitive blocks composed of convolution and non-linear operations as general CNNs. However, in a ResNeXt block, operations are performed across many branches and results are aggregated together with the block input (see \figurename ~\ref{resnext}(a)).
As ResNeXt is a 1000-categories ImageNet classifier, we substituted its last fully connected layer by some custom layers to make it a 2-categories classifier (see \figurename ~\ref{resnext}(b)). 

\subsubsection{Baseline:}
\paragraph{}
instead of learning from scratch, we started from a pretrained ResNeXt50 for each model of our hierarchical system. This is one way of doing transfer learning\cite{pretrained_surv} which is using a model trained for one task and re-targeting it for another related task. This is suggested in \cite{pretrained} as a baseline for any recognition task.

Then we fine tuned each model to its corresponding two classes. For the general model, we used the ImageNet\cite{imagenet} pretrained weights. For the two specialized models, we have chosen the best of: one fine tuned from ImageNet and  the other from the general model. The idea of fine tuning from the general model is that it has already seen these images and learned to extract meaningful features from them.

\subsection{Training}
\label{train}
We trained each CNN model of our system following these steps:
\begin{itemize}
  \item choosing an optimal learning rate $\eta$.
  \item training for 3 epochs the last layers randomly initialized using \cite{kaiming} while keeping pretrained ones fixed.
  \item training middle and last layers with different learning rates:
  \begin{itemize}
    \item $\eta/5$ for middle layers
    \item $\eta$ for last layers
  \end{itemize}
\end{itemize}
Additionally, during each training, we vary the optimal learning rate with a specific scheduling method. We will go over it and each step stated above with their explanations.

\subsubsection{Optimal learning rate choice:}
\label{lr}

the learning rate $\eta$ is one of the most important hyperparameter when training any CNN \cite{CycLearnRate}. In general it is set based on trial and error. We used a method in \cite{CycLearnRate} for setting an optimum value for $\eta$. The idea is to make one training run for few epochs while increasing $\eta$ from a very small value after each iteration. Then we plot the accuracy against the learning rate $\eta$ and note the value $\eta_{max}$ where the accuracy starts diverging or decreasing after increasing. The optimal $\eta$ is 1/3 or 1/4 of $\eta_{max}$. However we utilized an implementation of this method that uses the loss plot instead. We have found that in this case it works better when using 1/10 of $\eta_{max}$.

\subsubsection{Learning rate scheduling via Stochastic Gradient Descent with Warm Restarts (SGDR):}
\label{l_sgdr}
It is a method to schedule the learning rate variation during training so as to
converge rapidly. It has been proposed in \cite{sgdr}  and achieved new state-of-the-art results on CIFAR-10 and CIFAR-100 datasets.
In this approach, we decrease the optimal learning rate $\eta$ following the cosine annealing scheme until nearly zero. Then we suddenly set $\eta$ to its initial value and repeat again. This sudden jump of $\eta$ allows to look for another local minima around that may be better. That is the idea of "Warm Restarts".

\subsubsection{Training with different learning rates:}
\label{diff_lr}
This idea has been introduced by Jeremy Howard in \cite{diff_lr}. It is based on \cite{visualCNN} where authors show that first layers of CNNs learn to extract generic features like edges, corners, blobs and latter ones are more specialized on the task in hand. So we avoid to alter first layers as these features are useful for any task. In the same way, we slightly  alter middle layers because there are getting specialized on the task in hand and finally alter last layers with the optimal learning rate $\eta$ found. That is why we choose to use $\eta/5$ and $\eta$ respectively for middle layers and  last layers. The number is arbitrary and chosen based on trial and error but the idea of having decreasing learning rates from last to first layers remains. Here we set the learning rate of first layers to zero.

\section{Experiments and Results}

\subsection{Dataset}
\label{dataset}
The dataset is a set of images of hematoxylin and eosin (H\&E) stained breast histology microscopy. It is one of the two tracks of the ICIAR 2018 Grand Challenge on BreAst Cancer Histology (BACH) Challenge. It has 400 images equally distributed among four pathologies: \textit{Normal}, \textit{Benign}, \textit{In situ carcinoma} and \textit{Invasive carcinoma}. Images are conventional color images and are high resolution $2048\times 1536$  pixels with a pixel scale of $0.42 \mu m \times 0.42 \mu m$. This dataset is an extension of the BioImaging 2015 challenge\cite{dataext}.

We also used an additional public dataset from Bio-Image Semantic Query User Environment (BISQUE)\footnote{http://bioimage.ucsb.edu/research/bio-segmentation} \cite{bisque}. It is a very small dataset containing 58 histology images as the challenge but with less resolution: $896\times 768$ pixels. But the idea is to get a system that generalizes well so heterogeneous source of data is welcome. Images are distributed among two pathological groups: \textit{Benign} (32) and \textit{Malignant} (26).
These pathologies are excellent candidates for the two general pathological groups of the top model of our hierarchical system: \textit{Benign} belongs to Non carcinoma and \textit{Malignant} to Carcinoma.
But for the specialized CNN models, we used only images labeled as Benign for the one that classifies between Normal and Benign.

\subsection{Setup}
We trained our system on a Nvidia Tesla K80 equipped environment and used the following settings for training all three compounding CNNs:
\begin{itemize}
   \item splitting the dataset into 75\% for training and 25\% for validation.
   \item resizing all images to $ 299 \times 299 $. 
   \item augmenting the dataset by random rotations, reflections and cropping.
   \item setting batch size to 10.  
 \end{itemize} 

Concerning the resize operation, we first resized the image to $(299 \times ratio) \times 299$ which preserves the ratio equals to 4/3. So the resolution became $399\times 299$. Then we cropped a $299 \times 299$ square at the center of the resized image. This reduces of the risk of missing important parts of the image which are likely to be around its center. We did so to accelerate the training.

We implemented our system with a new library namely, fastai \cite{fastai} which is based on Pytorch  \cite{torch}, a framework for deep learning and GPU computations.

\subsection{Results based on the training set}
\label{results}

\begin{figure}[!t]
\centering
\includegraphics[width=2.5in]{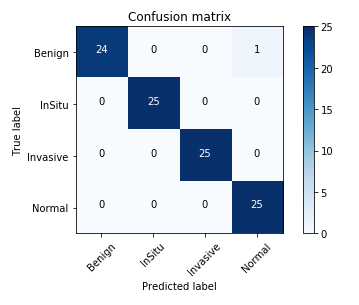}
\caption{Confusion matrix of the results on the BACH validation dataset.}
\label{cm}
\end{figure}

We evaluated our method on 25\% of the dataset unused during training. \tablename~\ref{res_tab} presents results. We performed  better on the initial dataset than the extended one. \figurename ~\ref{cm} shows the confusion matrix of results on the BACH dataset. Only one out of 100 images was misclassified.
We obtained these results in less than 60 epochs of training for each model which is very fast thanks to the training approach.

\begin{table}[t]
\caption{Performance of the method based on the validation set. 
Init. and Ext. are respectively BACH and BISQUE datasets.
Carci is the top model of the system. NorBe and InvIs are specialized models 
that classify respectively into Normal and Benign and into In situ and Invasive.
Last column is the preliminary competition test set result.}
\label{res_tab}
\begin{center}
\begin{tabular}{ |l|l|l|l|  }
 \hline 
 Models & Init. & Init. + Ext. & Competition\\
 \hline \hline
 Carci        & 1.00          & 0.98        & \textendash \\\hline
 NorBe        & 0.98          & 0.965       & \textendash  \\\hline
 InvIs        & 1.00          & \textendash & \textendash  \\\hline
 Whole system & \textbf{0.99} & 0.963       & \textbf{0.81} \\
 \hline
\end{tabular}
\end{center}
\end{table}

\subsection{Final system for the competition and result}

For the competition we assemble four different versions of the general or \textit{Carci} model and three versions of the specialized ones: \textit{NorBe} and \textit{InvIs} as named in \tablename~\ref{res_tab}.

\begin{enumerate}
   \item \textbf{\textit{Carci}}: two versions are in  \tablename~\ref{res_tab}. They correspond to \textquoteleft Init.\textquoteright~and \textquoteleft Init. + Ext.\textquoteright~columns. We have built the third version by training on the whole dataset (BACH and BISQUE combined). The last one is a snapshot of \textquoteleft Init\textquoteright column when accuracy was 0.99.
   
   \item \textbf{\textit{NorBe}}: similar to the first three versions of \textit{Carci} model. 
   
   \item \textbf{\textit{InvIs}}: \tablename~\ref{res_tab} contains one version which we obtained by using \textit{Carci}  as a pretrained model. The second version used the ImageNet pretrained model. Finally, we trained the model on the whole BACH dataset to build the last version.
 \end{enumerate} 
When training on the whole dataset, there is no way to check overfitting in contrast to having separate train and validation sets where losses of both help check it. Thus we trusted more these latter than those trained on the whole dataset. As reported in \tablename~\ref{res_tab}, this system reached an accuracy of 0.81 on the preliminary competition test set and rank us the 8$^{th}$ place out of 50.

\section{Related work and Discussions}
Computer-Aided Diagnosis (CAD) has become a major area of research in medical imaging included histology images\cite{cad}. There are many works on breast cancer detection from histology images with different datasets. It makes it difficult to fairly compare methods. Thus, competitions like BACH challenge are well suited for that and help advancing researches. 

In \cite{iciar_org}, a CNN based model has been developed to classify histology images in the same four classes as our problem. They used the BioImaging 2015 challenge\cite{dataext} dataset which is the basis of the BACH dataset. Our reported accuracy is better than them: 0.81 vs 0.778. But this does not imply our method is better since this dataset is larger. The best result on the preliminary results of the BACH challenge is an accuracy of 0.87 which not far from us.
Besides, looking at \tablename~\ref{res_tab}, we may suspect our system to overfit as there is a gap of 0.19 between validation and competition test set accuracies. To combat overfitting, we used dropout\cite{dropout} twice respectively with probabilities 0.25 and 0.5 (see \figurename ~\ref{resnext}(b)). Thus, we inclined towards our validation choice which may be similar to the training set. Additionally, mainly using 75\% of the data could have prevented our system to discover new patterns in validation data to perform better. Likewise, not only the resize operation induces lost of image parts but may also hamper the system to capture helpful details.

\section{Conclusion and Perspectives}
We proposed a hierarchical system of three CNN models to solve the image-wise classification of the BACH challenge. This system classifies gradually images into two categories carcinoma and non-carcinoma and then into the four classes of the challenge. When training CNN models, we followed a scheme that accelerate convergence. We got an accuracy of 0.81 on the competition test set and rank 8$^{th}$ out of 51 teams.
The first way to improve our system is training on the whole dataset using a strategy described in \cite{maxout}. It consists in training the whole dataset until the loss reaches that of the best accuracy obtained during validation. Regarding the resize operation, we can use a left and right crop with the center crop during training. Not only this will avoid losing part of images but serve as a data augmentation strategy as well. Finally, we could increase images' size but not exceeding the original size.

%
%
\bibliographystyle{splncs04}
\bibliography{iciararticle}

\end{document}